\documentclass{article}

\usepackage{arxiv}

\usepackage[utf8]{inputenc} 
\usepackage[T1]{fontenc}    
\usepackage{hyperref}       
\usepackage{url}            
\usepackage{booktabs}       
\usepackage{amsfonts}       
\usepackage{nicefrac}       
\usepackage{microtype}      
\usepackage{graphicx}
\usepackage{natbib}
\usepackage{doi}
\usepackage{tabularx}
\usepackage{adjustbox}
\usepackage{amsmath}
\usepackage{amssymb}
\usepackage{dutchcal}
\usepackage{float}
\usepackage{hhline}
\usepackage{mathtools}
\usepackage{multicol}
\usepackage{multirow}
\usepackage{wasysym}
\usepackage{geometry}
\usepackage{indentfirst}

\title{Towards the Development of an Uncertainty Quantification Protocol for the Natural Gas Industry}
\date{July 15, 2021} 

\author{ \href{https://orcid.org/0000-0003-4423-9833}{\includegraphics[scale=0.06]{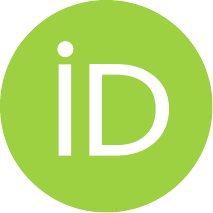}\hspace{1mm}Babajide ~Kolade}\thanks{Current email address: jide.kolade@fitila.ai} \\
	Institute Engineer\\
	Gas Technology Institute\\
	Des Plaines, IL. 60018 \\
	\texttt{jkolade@gti.energy} \\
}

\renewcommand{\headeright}{Technical Report}
\renewcommand{\undertitle}{Technical Report}
\renewcommand{\shorttitle}{UQ Protocol for Natural Gas Industry}

\hypersetup{
pdftitle={Towards the Development of an Uncertainty Quantification Protocol for the Natural Gas Industry},
pdfsubject={math-ph, CoRR},
pdfauthor={Babajide ~Kolade},
pdfkeywords={uncertainty quantification, modeling and simulation, machine learning, natural gas, vapor dispersion, pipeline risk assessment},
}

\begin{document}
\maketitle

\begin{abstract}
Simulations using machine learning (ML) models and mechanistic models are often run to inform decision-making processes. Uncertainty estimates of simulation results are critical to the decision-making process because simulation results of specific scenarios may have wide, but unspecified, confidence bounds that may impact subsequent analyses and decisions. The objective of this work is to develop a protocol to assess uncertainties in predictions of machine learning and mechanistic simulation models. The protocol will outline an uncertainty quantification workflow that may be used to establish credible bounds of predictability on computed quantities of interest and to assess model sufficiency. The protocol identifies key sources of uncertainties in machine learning and mechanistic modeling, defines applicable methods of uncertainty propagation for these sources, and includes statistically rational estimators for output uncertainties. The work applies the protocol to test cases relevant to the gas distribution industry and presents learnings from its application. The paper concludes with a brief discussion outlining a pathway to the wider adoption of uncertainty quantification within the industry.
\end{abstract}

\keywords{uncertainty quantification, modeling and simulation, machine learning, natural gas, vapor dispersion, pipeline risk assessment}

\section{Introduction}
Engineers and researchers often run simulations using machine learning models and mechanistic models to inform decision-making processes. For example, in a pipeline risk assessment exercise, a vapor dispersion model may be run to estimate distances between critical dispersant concentrations and identified consequence receptors. Or an image classification model may be run using satellite imagery to detect construction activity and alert pipeline operators of potential third-party damage. The results of such simulations are numeric and categorical values, and though they might be presented as precise, contain inherent uncertainties. Uncertainty estimates of simulation results are critical to the decision-making process because simulation results of specific scenarios may have wide, but unspecified, confidence bounds that may impact subsequent analyses and decisions. With respect to research and development functions, results of simulations run for exploratory analyses affect research direction and allocation of resources to downstream tasks. Promising solution pathways may be identified through simulation. These predicted pathways may have hidden pitfalls and unqualified probabilities of success that obscure their viability. Therefore, unquantified simulation results may impede the pace of scientific discovery and engineering analyses.

Machine learning (ML) models have had significant impacts on different industries in the last several years. These impacts have been aided due to the ease of collection and storage of data, ever-increasing computational power, and the recent development and success of powerful machine learning methods. However, some of the most widely adopted machine learning archetypes are `closed box' models that are inscrutable. The critical need to recognize hazards in complex natural gas distribution systems warrants models with auditable predictions. Given the success of some variants of ML models and the attention they have generated, it is expected that the pace of adoption of these models will accelerate. It is important that protocols and workflows that may be used to assess their suitability to inform decision-making processes be established now.

Uncertainty quantification (UQ) characterizes significant sources of uncertainties in a model and propagates their effects to computed quantities of interest (QOI)\footnote{In contrast, sensitivity analysis quantifies how the uncertainty in a specific source contributes to overall uncertainty in the computed QOI.}. Goals of uncertainty quantification may include sensitivity analysis, estimation of predictability bounds, evaluation of model sufficiency or deficiency, evaluation of model performance in abnormal environments, determination of reliability metrics, and the quantification of margins and uncertainties (QMU)\footnote{QMU focuses on margin-to-failure of a design compared to uncertainties in key computed QOIs.}. 

The objective here is to develop a protocol to assess uncertainties in predictions of machine learning and mechanistic simulation models. The protocol will outline an uncertainty quantification workflow that may be used to establish credible bounds of predictability on computed quantities of interest and to assess model sufficiency. The protocol will identify key sources of uncertainties in machine learning and mechanistic modeling, define applicable methods of uncertainty propagation for these sources, and include statistically rational estimators for output uncertainties. The protocol will present a concise and rigorous workflow that may be applied to a wide variety of models to assess their predictive capability. This protocol will help answer the question: “how do you know when to trust the result of a machine learning model?”

\subsection{Machine Learning and Uncertainty Quantification}
Machine learning models are templated mathematical models that are domain agnostic. That is, they are defined sequences of mathematical functions that may be used to model phenomena in different domains without necessarily appealing to the underlying governing laws: A defined \textit{template} of a sequence of mathematical functions may be used to model different classes of phenomena. For example, a recurrent neural network model template that is used for speech recognition may also be used for seismic activity detection. Machine learning algorithms derive parametric values for templated model structures from empirical data through their “learning” process\footnote{In statistical theory, this process is called \textit{inference}: the process of using empirical data to infer properties of the data generating mechanism.}. Ensuing models are used to infer unobservable quantities such the “true” system state or predict future observed states. Machine learning is an inverse problem. The resulting model parameterizations may not be stable under perturbation of the training dataset. Uncertainty quantification may be used to assess stability of parameterizations and variability of model predictions.

The rapid increase in the ease of data collection and cheap availability of data storage have spurred the adoption of machine learning models. Correspondingly, the scale and complexity of ML models have increased due to innovations in ML and statistical methods and the availability of heterogeneous computational hardware, such as graphics processing units (GPUs). UQ methods used to assess suitability of ML models must scale well with large data sets and work with complex models. 

Consumers of ML models are often non-experts in data science or statistics and are interested in applying such models to solve an item in their workflow domain of interest. The machine learning model is then a computational `closed box' and end users have no need or interest in understanding its inner workings. So, UQ protocols must be suitable in assessing the credibility of `closed box' models, without understanding the inner workings, and should be able to yield tighter uncertainty metrics for 'open box' models. The results from applying a UQ protocol to a model should be easily understood by non-experts. In summary, the following are desirable qualities of a UQ protocol applicable to ML models:

\begin{itemize}
	\item Performance scale well with large data

	\item Applicable to simple and complex models

	\item Work for a variety of data types and model types

	\item Work for high precision-low accuracy and low precision-high accuracy models

	\item Work when the model is a `closed box'

	\item Provide steps for tighter uncertainty metrics when model is an `open box' or `glass' box model
\end{itemize}

\subsection{Uncertainty Types}
Uncertainty classification used in this protocol is the taxonomy that is prevalent in the risk assessment and scientific computation communities \citep{PATECORNELL199695,Mehta.2016,ASME.2009}, which classifies uncertainties based on the knowledge of the state of a system and its parameters. Uncertainty sources are typically classified as aleatory, epistemic, or mixtures of both. This differs from the classification employed in some experimental and metrology protocols that classify uncertainty sources based on their presumed effect on the result or by the process with which they were quantified \citep{ASME.2006}. The choice of this taxonomy reflects the goal of the protocol: to outline a method with which credible bounds of the simulation results may be quantified based on the current state of knowledge.

Aleatory uncertainties are also referred to as stochastic or irreducible uncertainties. Aleatory uncertainties are uncertainties due to inherent spatiotemporal variations in a quantity or physical process. An example of aleatory uncertainty is the instantaneous wind velocity in a dispersion field. One may be able to bound wind speed and direction at a location of interest using weather models and historical data. However, the instantaneous wind velocity is completely stochastic. Aleatory uncertainties exhibit Markovian properties: additional information does not improve the quality of their prediction.

Epistemic uncertainties are uncertainty sources due to lack of knowledge about the system. Epistemic uncertainties are also referred to as reducible uncertainties as these uncertainties may be eliminated or reduced by additional experimental data, by higher-fidelity models, or by expert opinions. One useful method to distinguish these two types of uncertainties is: aleatory uncertainties are well characterized by probability distributions and epistemic uncertainties may be represented by categorical options or intervals. An example of a mixed form uncertainty is an uncertainty source that is well characterized by the Gaussian distribution but the mean and variance for that application is unknown.

\subsection{Approach}
Main contributions of this work to UQ literature are the 1) development of a unified framework with which uncertainties in predictions of mechanistic and machine learning models may be evaluated and 2) an in-depth exposition of factors that contribute to uncertainty in results of numerical models of complex natural gas distribution systems. Application of uncertainty quantification techniques to mechanistic models has a longer history and its workflows better developed than the application of the techniques to ML models. Here, we adapt methods applied to mechanistic simulation to ML models, present methods specific to ML, and present a concise and well-defined unified protocol. Though the protocol is motivated by modeling problems encountered in the natural gas distribution industry, it is expected that its applicability will have a wider audience. The novelty of this work is in streamlining UQ methods into a useful protocol for developers and consumers of numerical models applied in the natural gas industry.

The rest of the paper is laid out as follows: \nameref{sec:relatedWork} reviews prior work in the field of uncertainty quantification applied to both mechanistic modeling and machine learning. \nameref{sec:measuresAndMethods} discusses some of the mathematical underpinnings and algorithms applied in uncertainty quantification. This section is more theoretical than the rest of the text. But it is included to provide foundational support to the UQ workflow developed and discussed in the rest of the paper. The \nameref{sec:uqWorkflow} section concisely presents a unified UQ workflow applicable to mechanistic modeling and machine learning. The \nameref{sec:uqApplication} section applies the workflow to test cases and discusses learnings from its application. The paper concludes with the \nameref{subsec:integratedWorkflow} subsection, where a pathway to the wider adoption of UQ analyses within the industry is discussed.

\section{Related Work}
\label{sec:relatedWork}
\subsection{Related Work - Mechanistic Modeling}
Protocols developed by the American Society of Mechanical Engineers (ASME), such as the ASME Standard for Verification and Validation in Computational Fluid Dynamics and Heat Transfer \citep{ASME.2009}, and reports and articles published by Sandia National Laboratories \citep{Rider.2010,Oberkampf.2002,Oberkampf.2004,Oberkampf.2007,Helton.2009} have underpinned development of systematic work in the field of scientific computation verification, validation, and uncertainty quantification. Table \ref{tab:coreUQElements} summarizes core elements of common UQ workflow for mechanistic models.
\begin{table}
    \caption{Core elements of common UQ workflow}
    \label{tab:coreUQElements}
    \centering
    \begin{tabularx}{\textwidth}{lXX}
    \toprule
    Step     & Objective     & Method(s)       \\
    \midrule
    Source identification & Identify uncertainty sources that materially impact the QOIs for the stated UQ objective(s)  & Phenomena identification and ranking table (PIRT) \citep{Oberkampf.2004,Hills.2015}     \\
    \midrule
    Code verification     & Establish that the computer program solves the mathematical model as intended. That is, establish that there are no programming mistakes (“bugs”) in the program & Method of manufactured solutions (MMS) \citep{Oberkampf.2004,Hills.2015}      \\
    \midrule
    Solution verification    & Assess and quantify numerical errors induced through the numerical solution of partial differential equations.      & Grid refinement study, Grid Convergence Index via Richardson extrapolation \citep{Oberkampf.2004,Roy.2005,Jatale.2015}. \newline Finite element-based error estimators \citep{Zienkiewicz.1992a,Zienkiewicz.1992b}  \\
    \midrule
    Uncertainty propagation   & Propagate numeric values of uncertainty sources to computed quantities of interests.   & Sampling methods, e.g., Latin hypercube sampling \citep{ASME.2009, Helton.2003,Helton.2000,Yegnan.2002}. \newline Stochastic expansion, e.g., polynomial chaos expansion \citep{Iaccarino.2011,Mai.2016,Najm.2009,Doostan.2009}. \newline Optimization methods \citep{Mehta.2016,Bohnhoff.2019uf,Kochenderfer.2019}. \newline Reliability methods \citep{Bohnhoff.2019uf,Yegnan.2002,Swiler.2016}. \\
    \bottomrule
    \end{tabularx}
\end{table}

\subsection{Related Work - Machine Learning}
Uncertainty quantification of ML models has a more recent history than that of mechanistic models. The work of \citeauthor{Shrestha.2006} estimates prediction uncertainty from model error variances. They used fuzzy c-means to partition the input space into clusters having similar model errors. The work used a separation index that minimizes the total within-cluster to outside-of-cluster variation to determine the optimal number of clusters. Prediction intervals were calculated for each cluster. And prediction intervals were calculated for each sample point based on the clusters' values and its membership factors in the clusters. The calculated prediction intervals for sample points were used to create a “local uncertainty estimation model” that would be used to estimate prediction uncertainty for subsequent inputs.

Stracuzzi et al. \citep{Stracuzzi.2017,Stracuzzi.2018apw} applied uncertainty quantification techniques to multimodal image segmentation analysis. They used the bootstrap method \citep{Efron.1979} to generate multiple samples of optical and lidar images of the same scene. Machine learning models were trained on the bootstrap samples and used to evaluate likelihood probabilities. The probabilities were aggregated in the posterior and used to compute uncertainty metrics of model parameters. Their work demonstrates how UQ results may be used in a decision-theoretic framework. Particularly, they showed how to use Shannon entropy and Kullback-Leibler divergence to decide on the addition of a complementary data source. They compared predictive capability and uncertainty of predictions from using optical-only data channels to optical-and-lidar data channels. Their results show that optical-and-lidar was more predictive, but more uncertain, for the scenario considered. While this approach is well suited for propagating parametric uncertainties, it does not address data uncertainty.

\citeauthor{Loquercio.2020} addressed data uncertainty in their work, which focused on neural network predictions in robotics applications. Their method assumed data uncertainty to be normally distributed. Data uncertainty was propagated forward in the neural network using an approximation that assumes each layer's activation is normally distributed and independent. Their model is a stochastic activation neural network: They estimate model uncertainty through the variance of point estimates of the model from a MC simulation using a specified Bernoulli distribution to `drop out' model weights. The data uncertainty and model uncertainty are propagated independently; the total uncertainty is estimated as a sum of both components. Though these assumptions allow factorization of the joint probability and facilitate uncertainty propagation, they are restrictive given the nonlinearity afforded by neural nets and do not allow estimation of uncertainty due to the coupling effects of data and model.

Similarly, \citeauthor{Gal.2015} \citep{Gal.2015,Gal.2016} use stochastic activation neural networks to estimate parametric uncertainty. They use stochastic regularization techniques (SRTs), such as dropout, to inject stochasticity by displacing the effect of the SRT from the feature space to the parameter space. They use variational inference to estimate optimal Bernoulli parameters (Bernoulli variational distribution) by minimizing the Kullback-Leibler divergence between an approximate distribution and a Gaussian process model, with specified covariance structure, of the neural network. Stochastic activation neural networks, also called `Bayes via DropOut', are more efficient than full Bayesian neural networks (BNN) but do “not fully capture uncertainty associated with model predictions” \citep{Jospin.2020} and are not as computationally efficient \citep{Gal.2016} as bootstrap methods.

\section{Measures and Methods}
\label{sec:measuresAndMethods}
\subsection{Measures}
Uncertainty quantification is inextricably linked with probability theory. The goal of this section is to describe foundational aspects of probability and statistical theory that are applicable to the development of the UQ workflow.
\subsubsection{Probability Spaces}
The following is a brief description of measure-theoretic probability spaces \citep{Rosenthal.2006}. Let $\Omega$ be the set of possible outcomes of an experiment. For example, let an experiment be the tossing of a coin twice. Then, $\Omega$ is the set of possible outcomes, or the sample space, and $\omega$ are elements of the sample space. For this experiment,
\begin{center}
$\Omega = \{HH,HT,TT,TH\}$
\end{center}

Given the set $\Omega$, $\sigma$-algebra for $\Omega$ is a family, $\mathcal{F}$, of subsets of $\Omega$ with following properties:
 \begin{itemize}
     \item $\emptyset \in\mathcal{F}$: Null is a member of $\mathcal{F}$.
     \item $F\in\mathcal{F}\Longrightarrow F^c\in\mathcal{F}$: If $F$ is a member of the $\sigma$-algebra $\mathcal{F}$, then its complement $F^c$ is also a member.
     \item $F_1,\ F_2,\ldots\ \in\mathcal{F}\Longrightarrow F=\ \bigcup_{i=1}^{\infty}F_i \in\mathcal{F}$: If $ F_1$, $F_2$, and so on are members of $\mathcal{F}$, then their union, $F$, is also a member.
 \end{itemize}
For this experiment, one valid $\sigma$-algebra is: 

\begin{center}
$\mathcal{F}=\ \{\emptyset,\ \{HH,\ HT,\ TT,\ TH\},\ \{HH,\ HT\},\ \{TT,\ TH\}\}$
\end{center}
The pair $\left(\Omega,\mathcal{F}\right)$ is called a measurable space.
The probability measure, $\mathcal{P}$, maps the $\sigma$-algebra, $\mathcal{F}$, to the closed interval $\left[0,1\right], \mathcal{P}:\ \mathcal{F}\rightarrow\ \left[0,1\right]$, such that:
\begin{itemize}
\item $\mathcal{P}\left(\emptyset\right)=0,\ \mathcal{P}\left(\Omega\right)=1$
\item The probability of the union of disjoint outcomes is the sum of the probabilities of each outcome: $\mathcal{P}(\bigcup_{i=1}^{\infty}{F_i)=\ }\sum_{i=1}^{\infty}\mathcal{P}\left(F_i\right)$. 
\end{itemize}
The triple $\left(\Omega,\mathcal{F},\mathcal{P}\right)$ is called a probability space. A random variable, X, is the instantiation of a function that maps the sample space into the real space, $X:\Omega\rightarrow R^n$.

\subsubsection{Bayesian Probability}
Bayesian statistics is a powerful tool in uncertainty quantification. It is based on the Bayesian paradigm that probability is a measure of belief in the occurrence of events and not just the limit frequencies of events occurrences. And that, prior beliefs may be updated by incorporating new evidence. These notions are succinctly captured by Bayes' rule:
\begin{center}
\begin{equation}
\label{eq:bayesRule}
p\left(\theta\middle| D\right)=\ \frac{p\left(D\middle|\theta\right)p\left(\theta\right)}{p\left(D\right)}=\ \frac{p\left(D\middle|\theta\right)p\left(\theta\right)}{\int p\left(D,\theta^\prime\right)d\theta^\prime}
\end{equation}
\end{center}
In Bayesian statistics, $\theta$ is the hypothesis, such as the parameters in a machine learning model, and $D$, is the data, such as the training data. $p\left(D\middle|\theta\right)$ is the likelihood, $p\left(\theta\right)$ is the prior, and $p\left(D\right)$ is the evidence. Using the evidence, the posterior probability, $p\left(\theta\middle| D\right)$, is updated from the prior. The posterior gives a distribution of model parameters that can be used to calculate uncertainty of modeled QOIs. Though mathematically appealing, practical applications of the Bayesian paradigm to UQ are not common because of the lack of efficient algorithms. Methods to estimate the posterior probability, such as the Markov Chain Monte Carlo method, are discussed below.

\subsubsection{Kullback-Leibler Divergence and Jensen-Shannon Distance}
The Kullback-Leibler (KL) divergence is used to measure the discrepancy between two probability distributions. For uncertainty quantification and in Bayesian analysis, Kullback-Leibler divergence is often used to measure how far away a simpler distribution $q$ is from a more complicated distribution $p$: 
\begin{center}
\begin{equation}
\label{eq:klDiv}
KL\left(q\Vert p\right) = E_q\left[log\ q\left(x\right)\right]\ - E_q\left[log\ p\left(x\right)\right]
\end{equation}
\end{center}
where $E_q\left[\cdot\right]$ is the expectation of quantity in brackets over the distribution $q$. The base of the logarithm in the definition is often taken as $e$ or $2$. The measure is non-negative $(KL\left(q\Vert p\right)\ >\ 0)$, but it is unbounded, and it is not symmetric.

The Jensen-Shannon Distance is derived from the Kullback-Leibler divergence. The Jensen-Shannon Distance (JSD) is defined as:
\begin{center}
\begin{equation}
\label{eq:jsdist}
JSD\left(q\Vert p\right) = \sqrt{\frac{1}{2}KL\left(q\Vert M\right) + \frac{1}{2}KL\left(p\Vert M\right)}
\end{equation}
\end{center}
where $M=\frac{1}{2}\left(p+q\right)$. JSD is a symmetric and smooth version of the Kullback-Leibler divergence. Unlike the KL divergence, JSD is a distance metric and hence follows the triangle inequality. Using a logarithm base of 2, JSD is bound between 0 and 1: $0\leq JSD\left(q\Vert p\right)\leq1$. 

\subsubsection{Precision, Recall, F\textsubscript{1}}
Precision and recall are measures used to quantify global performance of machine learning algorithms. \textit{Precision} the ratio of positive cases that are correctly predicted. That is, the ratio of `true positive' to `predicted positive':
\begin{center}
\begin{equation}
\label{eq:precision}
precision=\frac{true\ positive}{true\ positive+false\ positive}
\end{equation}
\end{center}
Precision is also called the positive predictive value. \textit{Recall} is the ratio of `true positive' predictions to actual positive cases (`condition positive'):
\begin{center}
\begin{equation}
\label{eq:recall}
recall=\frac{true\ positive}{true\ positive+false\ negative}
\end{equation}
\end{center}
Recall is also called the `true positivity rate' or sensitivity. These measures are often not used in isolation; they are combined into the F-measure or the F\textsubscript{1} score via harmonic averaging:
\begin{center}
\begin{equation}
\label{eq:f1}
F_1=\frac{2\ast\ precision\ast r e c a l l}{precision+recall}
\end{equation}
\end{center}
These measures are widely used and are easy to use. However, they do not reward performance in correctly predicting negative cases \citep{Powers.2011}.

\subsection{Resampling Method:Bootstrap}
The bootstrap method \citep{Efron.1979} is used to estimate the distribution of a random variable that is a function of samples drawn for an unknown underlying population. For example, the outcome of an experiment run n times may be averaged to get its sample mean. The bootstrap method may be used to estimate the distribution and descriptive statistics (e.g., standard deviation) of the sample mean. The bootstrap method involves selecting n random samples, with replacement, to create B bootstrap samples. Therefore, a data point in the initial sample may be selected multiple times in a bootstrap sample. Each bootstrap sample may be thought as sampling from the underlying population in ``parallel universes.'' The number of elements in each of the bootstrap sample equals the number of elements in the initial sample. Then the function is evaluated for each bootstrap sample and results are collated from the sampling distribution and descriptive statistics are computed from the distribution. A sketch of the bootstrap algorithm is shown in Table \ref{tab:bootstrapAlgo}.
\begin{table}
    \caption{ Sketch of the bootstrap algorithm}
    \label{tab:bootstrapAlgo}
    \begin{adjustbox}{max width=\textwidth}
    \begin{tabular}{p{16.49cm}}
    \hline
    \multicolumn{1}{|p{16.49cm}|}{Bootstrap Algorithm} \\ 
    \hline
    \multicolumn{1}{|p{16.49cm}|}{Given a sample \textbf{X} with\ n\ observations and a function T }\\
    \multicolumn{1}{|p{16.49cm}|}{\(\mathbf{for }\ b = 1\ to\ B; \mathbf{do}\)}\\
    \multicolumn{1}{|p{16.49cm}|}{\textit{\hspace{6mm} $\#$Draw n samples, with replacement, uniformly at random for \textbf{X}.}}\\
    \multicolumn{1}{|p{16.49cm}|}{\textit{\hspace{8mm}\(\tilde{\mathbf{X}}^{(b)} = \{\tilde{X}_{i}^{\left(b\right)}\}\)}}\\
    \multicolumn{1}{|p{16.49cm}|}{\textit{\hspace{6mm} $\#$Evaluate function of bootstrap sample.}}\\
    \multicolumn{1}{|p{16.49cm}|}{\textit{\hspace{8mm}\( T_{b} = T(\tilde{\mathbf{X}}^{(b)})\)}}\\
    \multicolumn{1}{|p{16.49cm}|}{\(\mathbf{end for}\)}\\
    \multicolumn{1}{|p{16.49cm}|}{$\{$\( T_{b}, 1\leq b\leq B\)$\}$\( \leftarrow Sampling \ distribution\)} \\ 
    \hline
    \end{tabular}
    \end{adjustbox}
\end{table}

\subsection{Sensitivity Analysis: Morris-One-At-a-Time}
Morris-one-at-a-time (MOAT) is a screening method used to pare down the number of significantly sensitive inputs in a computational model. The method, originally proposed by \citeauthor{Morris.1991}, varies one input at a time in a randomized design to distinguish input variables that have either  negligible, linear and additive, or nonlinear/interaction effects on the output. Inputs with nonlinear or interactive effects on the output are selected for further analysis. Interested readers may see these references for additional information on MOAT sampling \citep{Saltelli.2007,Bohnhoff.2019uf}.

\subsection{Monte Carlo}
Monte Carlo methods are based on the direct application of the law of large numbers: That is, the empirical mean of a function whose inputs are independent and identically distributed (i.i.d.) samples converges (almost surely) to the expectation of the function as the sample size goes to infinity. Sampling methods such as the Latin hypercube sampling (LHS), importance sampling, and adaptive sampling use variance reduction techniques to significantly reduce the number of model evaluations for the same statistical precision. Importance sampling and adaptive sampling methods are generally more efficient than LHS \citep{Bohnhoff.2019uf}, but they may not be as easy to implement. Latin hypercube sampling ensures that the samples are representative of the underlying variabilities in the input space by selecting only one sample per hypercube \citep{Helton.2003}. Some implementations of the Latin hypercube sampling allow for incremental sampling and sampling from correlated inputs \citep{Bohnhoff.2019uf}.

\subsection{Stochastic Expansion}
\textit{Stochastic expansion} methods \citep{Mai.2016,Najm.2009} use probabilistic representations of the model's governing equations or system responses to propagate uncertainties in inputs to outputs. The model's uncertain parameters, or system responses, are expanded as finite sums of a set of orthogonal basis functions, reminiscent of the eigenfunction expansion method. With stochastic expansion, basis functions have random variables as inputs while their multiplicative coefficients (mode strengths) are deterministic functions. In the \textit{intrusive }variant, expansions of uncertain parameters are propagated through the governing equations using Galerkin projection to reformulate the model equation set in probabilistic terms. The adjective `intrusive' is because existing models would have to be rewritten to accommodate the reformulated model equations. \textit{Nonintrusive} stochastic expansion methods use model evaluations at (random or structured) sampled locations in the input space and orthogonality properties of the basis to calculate the mode strengths. The finite sum of the evaluated mode strengths multiplied by their corresponding probabilistic orthogonal basis function results in a probabilistic representation of the system response. As with the intrusive method, this probabilistic representation may be used to calculate full statistics of the system response. Though not as easily implementable, stochastic expansion methods typically converge faster than sampling methods for fewer number (e.g., fewer than ten) of uncertain inputs that are uncorrelated\citep{Roy.2011}.

\section{Uncertainty Quantification Workflow}
\label{sec:uqWorkflow}
The goal of this section is to present a well-defined protocol that may be used to assess uncertainties of mechanistic and machine learning models. This protocol builds on the ideas discussed in the previous sections and on well-established, peer-reviewed, guidelines. Though the protocol is intended to be general, the ensuing discussion uses specific examples to illustrate the workflow: 1) mechanistic modeling of gas dispersion from a leak in a natural gas pipeline and 2) image classification using a machine-learned model. It is assumed that the phenomena of interest may be represented by a model, $\mathfrak{M}$:
\begin{equation}
    \centering
    y = \mathfrak{M}\left(x;\theta,\varepsilon\right)
\end{equation}
For a continuous-space model, $y \in \mathcal{R}^m$ is the mode output, $x \in \mathcal{R}^n$ the model input, $\theta \in \mathcal{R}^p$, model parameters, and $\varepsilon \in \mathcal{R}^q$ the modeling uncertainties. The model transforms an n-dimensional\footnote{Note that multidimensional inputs may be reshaped as vectors.} input vector to an m-dimensional output ($\mathfrak{M}:\mathcal{R}^n \rightarrow \mathcal{R}^m$), given the model parameters and inherent uncertainties.

The following procedure is proposed as a workflow in assessing uncertainties in mechanistic and machine learning model computations used in assessing risks and predicting unobserved states in complex natural gas distribution systems:

\begin{enumerate}
    \item \textit{Define UQ objective(s) and define computed quantities of interest}: It is assumed that the model is of sufficient quality and has gone through software quality assurance (SQA) tests. SQA tests ensure the reliability of a model as software product that produces repeatable results on specified computer systems\citep{Oberkampf.2004}. The first step is to explicitly state the goal(s) of the UQ study and quantities of interest that the study will focus on.

    \item \textit{Identify sources of uncertainties}: Identify uncertainty sources that materially impact the quantities of interest for the stated UQ objective(s). Uncertainty sources are categorized as:

    \begin{enumerate}
        \item Input uncertainties

	\item Parametric uncertainties

	\item Numerical uncertainties

	\item Model-form uncertainties.

    These uncertainties can either be aleatory, epistemic, or mixed form uncertainties.

    \end{enumerate}
    \item \textit{Characterize uncertainties} by assigning a mathematical form (e.g., Gaussian random variable) to the uncertainty source and numerical values (e.g., mean, and standard deviation) to the form parameters. Resampling methods, such as the bootstrap method, may be needed to estimate numerical values for parametric uncertainties for machine learning models.

    \item \textit{Propagate uncertainties:} Propagate characterized uncertainty sources to outputs using problem appropriate UQ method (e.g., Latin hypercube sampling or stochastic expansion).

    \item \textit{Summarize UQ results} in a format best suited for the decision process. Formats include statistical metrics (e.g., standard deviations), confidence intervals, system response cumulative distribution functions, or probability boxes.

\end{enumerate}

The schematic below in Figure \ref{fig:imgUQWorkflow} summarizes the steps outlined for the UQ workflow. In practice, the UQ process is not as it is linearly depicted. Often when results of a UQ analyses are reviewed, the analyst may redefine the UQ objectives or expand the set of QOIs and repeat the process. The subsequent subsections go into detail describing each topic and then the workflow is applied to illustrative cases to present more concrete guidance.
\begin{figure}
   \centering
   \includegraphics[height=6in]{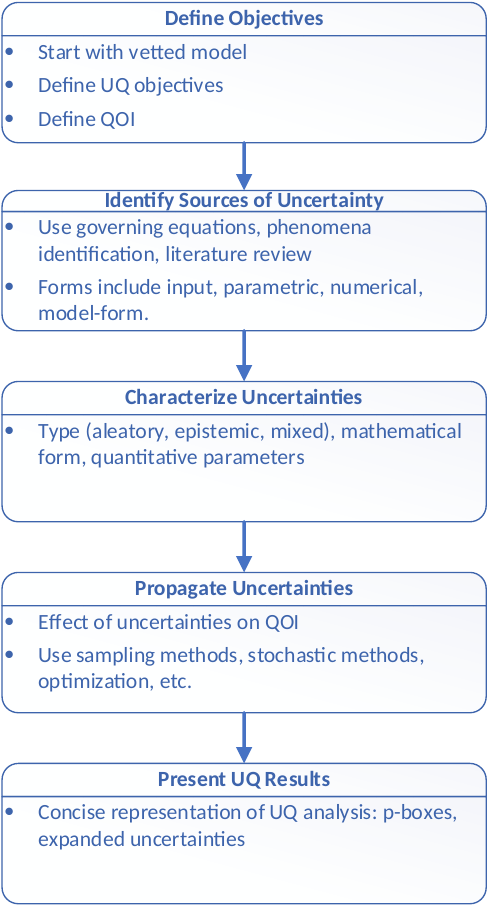}
   \caption{Summary of UQ Workflow} 
   \label{fig:imgUQWorkflow}
\end{figure}

\subsection{Sources of Uncertainties}
Sources of uncertainties in modeling and simulation include input uncertainties, parametric uncertainties, uncertainties due to numerical solution methods, and model-form uncertainties. 

\subsubsection{Input uncertainties}
\textbf{\textit{Input uncertainties}} are uncertainties in user-supplied information necessary to exercise the model to simulate the scenario of interest. These include uncertainties in descriptions of system geometry and layouts, initial conditions, boundary conditions, source terms, and forcing functions. Uncertainties in sensor measurements that are input to the model are also categorized as input uncertainties.

With respect to mechanistic modeling of contaminant dispersion, the \textit{initial system state and boundary conditions} to the model governing equations are significant sources of uncertainties in modeling and simulation. The location and geometry of plant equipment, walls, fences, and dikes are boundary conditions to the governing equations. Uncertainties about the numerical values associated with these inputs affect the quality of modeling predictions and decisions that can be informed by those predictions. Wind speed and atmospheric turbulence affect advection and turbulent entrainment and dispersion of contaminants. Uncertainties in \textit{source terms and forcing functions} are the most dominant source of uncertainties in dispersion modeling. The rate and duration of a leak dictate the overall mass of contaminants released to the atmosphere. At a minimum, an uncertainty quantification workflow must assess effects of these uncertainties on the spatiotemporal concentration profiles and other predicted quantities of interest.

Inputs to machine learning models are derived from field measurements and outputs from sensors. For example, machine learning models trained to predict pipeline corrosion rates may use predictors such as pipeline burial depth, soil composition, soil pH and conductivity, and terrain drainage, and other factors\citep{Ricker.2010}. These inputs, often in the form voltage signals, images, video segments, or other modalities represented as four-dimensional data sets, are susceptible to sensor noise, sensor failure, transcription errors, and data corruption. Uncertainty estimates in these input values should be propagated to the predicted quantity of interest.

\subsubsection{Parametric uncertainties}
\textbf{\textit{Parametric uncertainties}} are uncertainties with respect to values assigned to model parameters or ``model constants." In mechanistic modeling, model parameters are derived from simplifying assumptions, constitutive relationships, material properties, and calibrated to prior experimental or historical data. With respect to fluid dispersion modeling, these uncertainties may be due to the intrinsic variability in physical properties of the fluid or with model constants used in phenomenological models. Thermophysical properties of fluid components are well characterized and appropriate numerical values with respect to system properties are often well defined\citep{Linstrom_Nist.2001,Zarr.2009}. However, phenomenological model constants are often calibrated to historical data\citep{Rider.2010} and the analyst may lack the knowledge on what optimal values should be.

Parametric uncertainty is a significant source of uncertainty in machine learning models. Machine learning algorithms derive model parameter values from empirical data that is assumed to be a representative sample of the underlying distribution. The derived parametrization may be unstable under perturbation of the training dataset. Additionally, the test sample may belong to a region of the sample space that is not well represented by the training set, and hence predictions have wide uncertainties. Machine learning models are often `closed box' models and users have no access to parameter values. In such scenarios, effects of parametric uncertainties may go unassessed. When there is access to the ML model training workflow and dataset, one may use resampling methods or Bayesian methods to quantify the distribution of the model's parametric values. 

\subsubsection{Numerical uncertainties}
\textit{\textbf{Numerical uncertainties}}. The governing equations for most mechanistic and machine learning models do not admit closed-form analytical solutions. \textit{Numerical} approximations to the governing equations induce uncertainties in the computed solution. These uncertainties are introduced at various stages of the computation: 

\begin{enumerate}
    \item through the conversion of governing equations to discrete analogues \textemdash \textit{discretization error}; 

    \item in the representation of real numbers by the finite-precision arithmetic of floating-point numbers \textemdash \textit{round-off error};

    \item through numerical errors incurred when using approximate iterative solution methods \textemdash \textit{convergence error}; 

    \item through mistakes in programming the computer model \textemdash \textit{coding error; }

    \item and through approximation errors incurred when surrogate models, such as response surface methodology or Gaussian process models, are used to accelerate model evaluations as part of the UQ workflow.
\end{enumerate}

Solution and code verification methodologies encompass the formal methods for quantifying numerical errors \citep{ASME.2009,Roache.2002,Rider.2010}. Sandia National Laboratories, NASA, and the ASME have published a significant body of work on quantifying these uncertainties\citep{Mehta.2016,ASME.2009,Rider.2010,Oberkampf.2002}. These references contain cookbook-style workflows for quantifying numerical uncertainties. Discretization errors are often the largest source of numerical uncertainty\citep{Mehta.2016,Roy.2011,Roy.2005}. Application of Richardson extrapolation is the most popular method of quantifying uncertainties due to discretization error because it is non-intrusive and is applicable to finite-difference, finite-volume, and finite-element discretizations\citep{Roy.2005}. Numerical uncertainty should be treated as epistemic uncertainty because it is reducible: increase in mesh quality, adoption of a more accurate numerical method, or increase in computational resources may reduce this uncertainty.

Machine learning models do not typically calculate quantities of interest at discrete temporal and spatial points and so would not incur discretization errors. However, these models run on GPUs that use single-precision and `half'-precision floating-point architecture and so have increased numerical uncertainty due to round-off error. Additional numerical uncertainty is introduced to machine learning model results through the iterative optimization algorithms (e.g., stochastic gradient descent) used to induce model parameters. Since ML models use templated functions, coding errors in ML models will be displaced into the parameter values and will affect model interpretability. The errors discussed in this subsection represent uncertainty sources that should be considered as well as characterized and propagated to computed quantities of interest, if warranted.

\subsubsection{Model-form}
\textit{Model-form} uncertainties pertain to the choice of governing equations, abstractions, approximations, and mathematical form used to represent the physical system. There is often a hierarchy of models that may be used to represent a physical phenomenon. These models may be ordered in terms of their complexity, fidelity, or computational cost. For example, in modeling turbulent dispersion of contaminants, direct numerical simulation (DNS) is the most accurate and most computationally demanding method\citep{Pope.2000}. Large-eddy simulation (LES), PDF methods, and turbulent-viscosity models (e.g., k-$\varepsilon$ model) are not as accurate, in that order, as DNS but have progressively lower computational cost. Similarly, on the ML front, models used to predict sequences, such as attention-based networks\citep{Vaswani.2017}, long short-term memory (LSTM) networks\citep{Graves.2014}, gated recurrent units\citep{Chung.2014}, and recurrent neural networks\citep{Sherstinsky.2018}, respectively have decreasing computational demands and predictive performances. 

Theoretical model developers make simplifying assumptions to underlying mechanistic equations or conceptual framework to produce computationally tractable forms applicable only to specific scenarios of the phenomenon. These forms are often `good enough' for intended scenarios but may incur significant errors when applied outside their range of applicability, resolution capability, or when inputs violate simplifying assumptions. These errors are model-form uncertainties and are characterized through the validation process: comparing model results against experimental data or results of high-fidelity simulations. Model-form uncertainties should be treated as epistemic uncertainties because higher-fidelity models and more/higher-quality experimental data may reduce this uncertainty. 

\subsection{Characterizing Uncertainties}
Characterizing uncertainties includes assessing sources to determine whether they would contribute significantly to uncertainties in the computed quantities of interest. Computational resources limit the number of variables that can be considered in detail in the uncertainty propagation step. A screening method such as the Morris one-at-a-time\citep{Morris.1991} method may be used to preselect a smaller pool of variables to be considered in detail. It should be documented if a source is judged non-impactful and the reasons explicitly stated. For uncertainty sources that impact computed quantities of interest, characterization involves specifying the uncertainty type, specifying a mathematical model for the uncertainty, and specifying numerical values for the model parameters\citep{Roy.2011}.

Mathematical models/frameworks for representing uncertainties include probability theory, interval analysis, fuzzy set theory, possibility theory, and evidence theory\citep{Helton.2009}. Aleatory uncertainties are characterized by probability theory using probability density or cumulative distribution functions. For purely aleatory uncertainties, the parameters of the distribution function would be well defined. Mixed aleatory-epistemic uncertainties may be characterized by specified probability distributions but with uncertain parameters. For example, the wind speed profile used in a dispersion simulation may be characterized using the Weibull distribution with the parameters of the distribution characterized as epistemic uncertainties. Additionally, covariance of correlated uncertain inputs should be considered when characterizing aleatory uncertainties.

Interval analysis assumes that appropriate values for an uncertain parameter exist in a specified closed and bounded interval. However, no additional information is given about the `likelihood' of the values within that interval. It is not assumed that values within that interval have equal `likelihood', but that they are possible. Optimization methods may be used to derive extrema of quantities of interest, subject to the bounds on the uncertain input values\citep{Bohnhoff.2019uf}.

Evidence theory, also called Dempster-Shafer theory of evidence, uses two complementary measures to quantify `likelihood' of uncertain parameters: Belief and Plausibility. The `Belief' function measures the positive evidence supporting the validity of a proposition while the `Plausibility' function measures the absence of evidence refuting the proposition\citep{Helton.2009,Tucker.2006}. Combining both functions yields lower and upper limits for the probability of uncertain parameters, where the `Belief' function is lower than or equal to the `Plausibility' function. Quantitatively, uncertain parameters are represented by multiple intervals with associated basic probability assignments (BPAs). The intervals may overlap or be disjointed. The BPAs for a parameter must sum up to one and they quantify the likelihood of an uncertain parameter falling within an interval.  As with interval analysis, no additional information is given, or assumption made about the `likelihood' structure within the interval. Evidence theory adds a higher level of sophistication to quantifying epistemic variables than interval analysis and is more amenable to pooling degrees of beliefs polled from subject matter experts. 

\subsubsection{Machine Learning}
Parametric uncertainties of machine learning models are characterized using the training dataset typically employing one of two methods: The bootstrap method\citep{Efron.1979} or stochastic regularization techniques\citep{Hinton.2012}. Resampling methods such as the bootstrap method creates bootstrap samples of training dataset and retrains the model on each sample. Optimal parameter values derived from using each of the bootstrap samples are collated and descriptive statistics of the collection are used to characterize parametric uncertainties. In the SRT approach, the machine learning model is often modified by overlaying Bernoulli-type logic gates on each parameter to pass through or to zero-out that parameter value for an instance of the training data. Variational inference\citep{Hoffman.2013,Tsilifis.2016} or MCMC methods\citep{Craiu.2014,Hoffman.2014,Haugh.2017} are used to derive the posterior distribution of the parameters and hence characterize parametric uncertainties. The Bootstrap method is more computationally-efficient\citep{Gal.2016} and easier to apply for a wider range of ML models used by non-experts.

\subsection{Uncertainty Propagation}
Methods used to propagate numeric values of uncertainty sources to computed QOIs may be classified as sampling methods\citep{ASME.2009,Helton.2003,Helton.2000,Yegnan.2002}, stochastic expansion methods\citep{Mai.2016,Najm.2009,Doostan.2009}, optimization methods\citep{Mehta.2016,Bohnhoff.2019uf}, and reliability methods\citep{Mehta.2016,Bohnhoff.2019uf,Yegnan.2002}. \textit{Sampling methods} are the most widely used methods for uncertainty propagation. Sampling methods are applicable to aleatory uncertainties where the structure of the input sample space is probabilistically well characterized. The \textit{Monte Carlo} method uses random sampling of the input space, using specified distribution functions, to create simulation samples for the model. The model is evaluated at the specified samples and statistical estimations of the computed QOI are developed from the simulation results. The Monte Carlo method is easy to understand and to implement. As with other sampling methods, MC sampling is applicable to `closed box' models without any dependence on the underlying physics. The main disadvantage of Monte Carlo sampling is that it requires a large number of model evaluations to produce converged statistics. 

\subsection{Representation of UQ Results}
\begin{figure}
    \centering
    \includegraphics[width=4.00in]{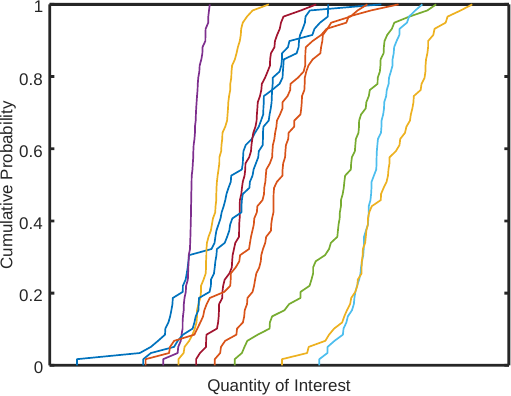}
    \caption{An example of a CDF ensemble plot used to summarize UQ results.}
    \label{fig:imgCDFExample}
\end{figure}

Results of the UQ analysis may be represented by an ensemble plot of cumulative distribution functions (CDFs) of the QOI to summarize the system's response to the uncertain parameters.  The CDF ensemble plot, sometimes called a `horsetail' plot, summarizes the system's response to aleatory variables in a single CDF curve for different instantiations of epistemic variables. Figure \ref{fig:imgCDFExample} shows an example of a `horsetail' plot. From such plots, one can read off the interval range of probabilities (`y-axis') for a specified system response (`x-axis') and the interval range of predicted QOI at specified probabilities. The outline of the `horsetail' plot is denoted as the probability box, or p-box. The results of the UQ analysis may also be summarized by expanded uncertainty formats by specifying the measurand and a coverage factor (hence confidence interval). For classification models, results of the UQ analysis may be summarized in PDF or histogram plots of the system's response to different input categories.

\section{Application of Uncertainty Quantification Workflow}
\label{sec:uqApplication}

\subsection{Dispersion Modeling}
In this section, we apply the UQ methodology discussed above to the case of contaminant dispersion modeling. 

\subsubsection{Define Objectives}
The objective of the UQ analysis is to define a methodology to put credible bounds of predictability on computed QOIs. In dispersion modeling, the primary QOI is the maximum distance to the lower flammability limit (LFL) of a combustible dispersant or distance between critical dispersant concentrations and identified consequence receptors. 

\subsubsection{Identify Sources of Uncertainty}
For a reasonably-vetted model, the UQ workflow, as described above, starts with identifying the sources of uncertainty that materially impact the computed QOI. Natural gas dispersion from a leak or rupture in a pipeline system may be modeled by non-isothermal, multiphase, multicomponent turbulent transport equations, including the continuity equation, momentum balance, energy conservation equations, and accompanying turbulence modeling equations. Due to computational cost, the generalized equations are adapted with simplifying assumptions. In fields with limited exposition, one may start the identification process by enumerating all the variables and parameters in the applied governing equations and use sensitivity analysis to narrow down the list of most impactful variables. However, extensive open literature on dispersion modeling and dispersion experiments gives a more concise starting point for a UQ study. Review of open literature indicates that the most impactful variables in UQ of dispersion modeling are the source term, wind advection, buoyancy effects, and turbulent mixing.

\citeauthor{Bedwell.2018} conducted a comprehensive literature review of atmospheric dispersion models to determine the most influential variables that contribute to the uncertainty of dispersion modeling predictions. Their work, sponsored by the European Joint Programme for the Integration of Radiation Protection Research, focused on dispersion of accidental radiological release, but also considered non-radiological dispersion modeling. Their review started with over 100 documents in which both Gaussian models and computational fluid dynamics (CFD) simulations were used. They used different criteria to judge importance of the variables, including using a raw count of the appearance of the variable in studies; assigning points based on relevance attached to variables by original authors; and categorization based on modeling approach, relevance of original paper, or groupings of variables. Source term, wind direction, wind speed, plume properties, and turbulent diffusion parameters consistently ranked highest. 

\citeauthor{Koopman.2007} and \citeauthor{LuketaHanlin.2006a} reviewed past work on LNG safety research and the large-scale experiments on the release and dispersion of LNG. The reviewers noted that the effect of turbulence or the lack thereof, due to stable density stratification, was key to dispersion distance and its prediction. This suggests that wind speed and atmospheric turbulence (through turbulence intensity, or stability class, mixing height, vertical diffusion parameters, and horizontal diffusion parameters) should be included in a UQ study.

\subsubsection{Characterize Uncertainties}
Given the preceding discussions, the following parameters are suggested to be included in uncertainty quantification for a CFD-based or integral-type model. The UQ analyst should select specific parameters equivalent to those listed in Table \ref{tab:uqSourcesCFDDispersion} for other model types based on the preceding discussions.
\begin{table}[H]
\caption{Identification and characterization of uncertainty sources for CFD-based dispersion model}
\label{tab:uqSourcesCFDDispersion}
\begin{adjustbox}{max width=\textwidth}
\begin{tabular}{p{4.91cm}p{2.23cm}p{2.86cm}p{6.53cm}}
\hline
\multicolumn{1}{|p{4.91cm}}{\centering
\textbf{Uncertain Parameter}} & 
\multicolumn{1}{|p{2.23cm}}{\centering
\textbf{Category}} & 
\multicolumn{1}{|p{2.86cm}}{\centering
\textbf{Classification}} & 
\multicolumn{1}{|p{6.53cm}|}{\centering
\textbf{Characterization}} \\ 
\hline
\multicolumn{1}{|p{4.91cm}}{\raggedright
Inlet mean wind speed (\(\overline{U}\))} & 
\multicolumn{1}{|p{2.23cm}}{Boundary \newline condition} & 
\multicolumn{1}{|p{2.86cm}}{Aleatory} & 
\multicolumn{1}{|p{6.53cm}|}{\(\overline{U }\sim  Weibull\left(2.0 m/s, 1.5\right)\)} \\ 
\hline
\multicolumn{1}{|p{4.91cm}}{\raggedright
Turbulence intensity (\( TI\))} & 
\multicolumn{1}{|p{2.23cm}}{Boundary \newline condition} & 
\multicolumn{1}{|p{2.86cm}}{Aleatory} & 
\multicolumn{1}{|p{6.53cm}|}{\centering
Z\(  \sim  TruncatedNormal\left(0,1,-2.7,\infty\right)\) \\ \( TI = 0.12\ast\left[ 0.75+\frac{\left(3.8+1.4Z\right) }{\overline{U }}\right]\)} \\ 
\hline
\multicolumn{1}{|p{4.91cm}}{Monin-Obukhov length (L)} & 
\multicolumn{1}{|p{2.23cm}}{Parameter} & 
\multicolumn{1}{|p{2.86cm}}{Epistemic} & 
\multicolumn{1}{|p{6.53cm}|}{\(\left\{ L | 10 m\leq L\leq 50 m\right\}\)} \\ 
\hline
\multicolumn{1}{|p{4.91cm}}{Discretization error} & 
\multicolumn{1}{|p{2.23cm}}{Numerical} & 
\multicolumn{1}{|p{2.86cm}}{Epistemic} & 
\multicolumn{1}{|p{6.53cm}|}{\( Use\ GCI\ to\ characterize\)} \\ 
\hline
\end{tabular}
\end{adjustbox}
\end{table}

This protocol will be applied in assessing predictive capabilities of dispersion models, often in comparison to one another. Hence, the phenomenological parameters of specific models are not considered here. Also, the source term which is a significant source of uncertainty in dispersion modeling is not included here because it may not be a selective parameter when comparing predictive capabilities of models. The source term should be included in UQ studies of specific scenarios. 

\begin{figure}
\centering
\includegraphics[width=6.5in]{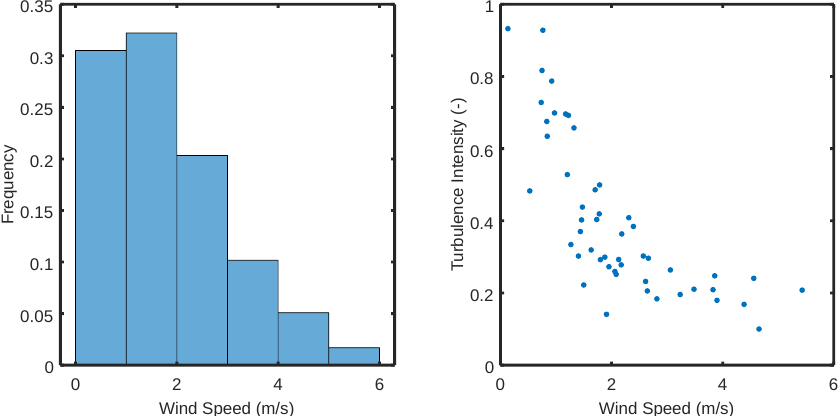}
\caption{Distribution of 59 random samples of wind speed and turbulence intensity using the UQ characterization functions specified in Table \ref{tab:uqSourcesCFDDispersion}}
\label{fig:windSpeedTI}
\end{figure}

The mean wind speed is specified at the inlet of CFD computational domain. The two-parameter Weibull function is a good approximation of atmospheric wind speed distribution\citep{Justus.1978}. The parameters listed in Table \ref{tab:uqSourcesCFDDispersion} result in a low wind speed with an average speed of 1.8 m/s, similar to the Burro 8 tests\citep{LuketaHanlin.2006a}. The low speed enhances gravity flow of dispersants and results in longer dispersion lengths, and possibly a more challenging modeling scenario. Wind turbulence intensity is correlated to wind speed and the correlation would complicate the random sampling process. The use of the dummy variable `\( Z\)' in the characterization above removes the correlation in the sampled space. Figure \ref{fig:windSpeedTI} shows random samples of wind speed and turbulence intensity using the UQ characterization functions specified above.

The shape of the atmospheric velocity profile is determined by the Monin-Obukhov length scale and accompanying universal functions. The length scale is partly determined by the surface heat flux. The inclusion of the length scale in the uncertain parameters reflects the uncertainty around the surface heat flux. The length scale specifies the location where the turbulence generated by buoyancy exceeds the turbulence generated by wind shear, and it is also an indication of atmospheric stability. The range of length scale specified in Table \ref{tab:uqSourcesCFDDispersion} corresponds to slightly stable atmospheric conditions. 

As discussed earlier, discretization error is often the largest source of numerical uncertainty for mechanistic models. The use of the grid convergence index is accepted as a standard in quantifying numerical uncertainties due to discretization errors\citep{Roache.2009}. GCI, based on Richardson extrapolation, uses the assumption that discretization error can be related to the grid size through a power series expansion. That is: \( E_{h} = Ch^{p}+H.O.T\). Where, \( h\) is the nominal grid size for the solution mesh, \( E_{h}\) is the discretization error at that grid size,\(  p\)  is the order of convergence, \( C\) is a proportionality constant, and \( H.O.T\). are higher order terms. The method uses computation of an integral QOI, here the maximum distance to LFL, at three different grid sizes to compute the order of convergence and estimate the numerical error; a `factor of safety' is multiplied with the numerical error estimate to the numerical uncertainty (\( U_{Num}\)). The ASME Standard\citep{ASME.2009} includes a concise five-step procedure for applying this method in a UQ workflow for CFD models. Interested readers and model evaluators should consult that standard or the book by \citeauthor{Roache.2009} for detailed steps.

\subsubsection{Propagate Uncertainties}
Uncertainties in the mean wind speed, inlet turbulent intensity, and Monin-Obukhov length are propagated to the QOI through model simulation. The numerical uncertainty, characterized using the GCI, is included in the overall uncertainty metric subsequently. The uncertainty set includes both aleatory and epistemic uncertainties. These uncertainties may be propagated to the computed QOI using a nested propagation approach, where the inner loop of the UQ propagation contains the aleatory sources and the outer loop contains the epistemic sources.

It is suggested that LHS be used to generate samples for the set of aleatory uncertainties, the inner loop, and propagated through the model to the computation of the QOI. The number of LHS samples needed to yield converged statistics in the computed QOI is not known \textit{a priori.} Given the potential for burdensome computation cost, the analyst should employ a progressive sampling procedure: start with a moderate LHS sample size and add samples till convergence is reached. The criterion for LHS convergence may be defined by Equations \ref{eq:QOIExpectationConv} or \ref{eq:QOIVarConv}:
\begin{center}
\begin{equation}
\label{eq:QOIExpectationConv}
\left\vert\frac{E\left[Q^{\left(i\right)}\right] - E\left[Q^{\left(i-1\right)}\right] }{E\left[Q^{\left(i-1\right)}\right]}\right\vert \leq \epsilon_{Q} 
\end{equation}
\end{center}

\begin{center}
\begin{equation}
\label{eq:QOIVarConv}
\frac{Var\left[Q^{\left(i-1\right)}\right] - Var\left[Q^{\left(i\right)}\right]}{Var\left[Q^{\left(i-1\right)}\right]}\leq \epsilon_{Var\left(Q\right)}
\end{equation}
\end{center}

where, \( E\left[Q^{\left(i\right)}\right]\) is the mean of the QOI from the LHS simulations using the current sample size, \( E\left[Q^{\left(i-1\right)}\right]\) is the mean of the QOI from the previous sampling iteration using a smaller sample size, and \( \epsilon_{Q}\) and \( \epsilon_{Var\left(Q\right)}\) are relative tolerance thresholds for criteria using mean value and variances of the QOI, respectively. A sample size of 59 is often used in Monte Carlo simulation of loss-of-coolant accident (LOCA) performance analysis in the nuclear industry\citep{Zhang.2016}.  The size is based on Wilks' formula for 95$\%$ confidence level on a single QOI prediction\footnote{ To streamline the workflow, the UQ analyst may choose to fix the size of the Monte Carlo samples to 59 without employing a progressive sampling procedure.}. \citeauthor{Shields.2015} in their study of progressive sampling methods used 0.01 for their relative tolerance thresholds. Figure 4 shows a depiction of the sampling procedure. Statistical software and numerical platforms commonly used to perform UQ analyses contain algorithms and strategies for such progressive sampling procedure\citep{Bohnhoff.2019uf,Gel.2013}.
\begin{figure}
    \centering
    \includegraphics{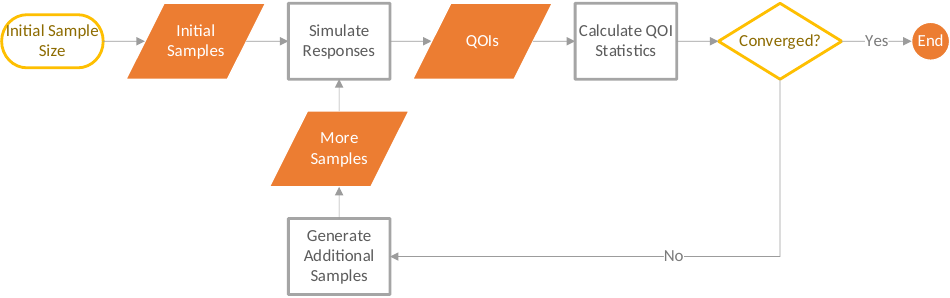}
    \caption{Progressive sampling procedure for aleatory uncertainties}
    \label{fig:progSamplingProc}
\end{figure}

In this mixed aleatory-epistemic UQ workflow, the outer loop epistemic uncertainty, the Monin-Obukhov length, is characterized by an interval. And no additional probability structure is specified about the interval. Samples may be generated for the outer loop using a design of experiments combinatorics approach. Then for a fixed outer loop sample, propagation of uncertainties in the inner loop proceeds as described above. An ensemble plot of the cumulative distribution functions (e.g., \ref{fig:imgCDFExample} for fixed outer loop samples may be used to define bounds of predictability of the QOI at different percentiles. Note that no one CDF has a higher or lower probability of occurrence than another CDF since they stem from interval-valued epistemic uncertainties. 

Instead of sampling the epistemic space, it may be more computationally efficient to treat outer loop as a global optimization problem. Since the protocol is concerned with defining bounds of predictability (minima and maxima), a global optimization algorithm can be used to find extrema in the epistemic sample space and compute the predictability bounds directly.

\subsubsection{Representation of Results}
The results of the UQ analysis should be represented by a probability box plot, which shows the minima and maxima of the CDF ensemble. Additionally, the expected value of the QOI and its expanded uncertainty should be reported for both bounding curves of CDF ensemble plot. The numerical uncertainty should be included in this calculation. The expectation of the QOI and its expanded uncertainty may be calculated as:
\begin{center}
\begin{equation}
    Q\in\left[\overline{Q}^{\left(LB\right)},\overline{Q}^{\left(UB\right)}\right]
\end{equation}
\begin{equation}
    (\overline{Q}^{\left(LB\right)} = min\left(E\left[\mathbf{Q}\right]-k\mathbf{\sigma }_{Q}\right)-U_{Num}
\end{equation}
\begin{equation}
    \overline{Q}^{\left(UB\right)} = max\left(E\left[\mathbf{Q}\right]+k\mathbf{\sigma }_{Q}\right)+U_{Num}
\end{equation}
\end{center}
where \( Q\) is the computed QOI, \(\mathbf{Q}\)\textbf{ }is the set CDFs corresponding to different instantiations of epistemic variables, k is the coverage factor for a specific confidence level (e.g., k$\sim$2 for 95$\%$ confidence level), and \( U_{Num}\) is the numerical uncertainty.

\begin{table}
\caption{Summary of Dispersion UQ Workflow}
\label{tab:uqWorkflowDispersion}
\begin{adjustbox}{max width=\textwidth}
\begin{tabular}{p{1.22cm}p{4.61cm}p{2.11cm}p{2.7cm}p{6.34cm}}
\hline
\multicolumn{5}{|p{16.98cm}|}{Define QOI} \\ 
\hline
\multicolumn{1}{|p{1.22cm}}{} & 
\multicolumn{4}{|p{15.76cm}|}{\raggedright
Maximum distance to LFL} \\ 
\hline
\multicolumn{5}{|p{16.98cm}|}{Identify Sources of Uncertainties} \\ 
\hline
\multicolumn{1}{|p{1.22cm}}{} & 
\multicolumn{4}{|p{15.76cm}|}{\raggedright
Source term, wind advection, buoyancy effects, and turbulent mixing.} \\ 
\hline
\multicolumn{5}{|p{16.98cm}|}{Characterize Uncertainties} \\ 
\hline
\multicolumn{1}{|p{1.22cm}}{} & 
\multicolumn{1}{|p{4.61cm}}{\centering
\textbf{Uncertain Parameter}} & 
\multicolumn{1}{|p{2.11cm}}{\centering
\textbf{Category}} & 
\multicolumn{1}{|p{2.7cm}}{\centering
\textbf{Classification}} & 
\multicolumn{1}{|p{6.34cm}|}{\centering
\textbf{Characterization}} \\ 
\hline
\multicolumn{1}{|p{1.22cm}}{} & 
\multicolumn{1}{|p{4.61cm}}{\raggedright
Inlet mean wind speed (\(\overline{U}\))} & 
\multicolumn{1}{|p{2.11cm}}{Input} & 
\multicolumn{1}{|p{2.7cm}}{Aleatory} & 
\multicolumn{1}{|p{6.34cm}|}{\(\overline{U }\sim  Weibull\left(2.0 m/s, 1.5\right)\)} \\ 
\hline
\multicolumn{1}{|p{1.22cm}}{} & 
\multicolumn{1}{|p{4.61cm}}{\raggedright
Turbulence intensity (\( TI\))} & 
\multicolumn{1}{|p{2.11cm}}{Input} & 
\multicolumn{1}{|p{2.7cm}}{Aleatory} & 
\multicolumn{1}{|p{6.34cm}|}{\centering
Z\( \cdot\sim \cdot TruncatedNormal\left(0,1,-2.7,\infty\right)\) \\ \( TI = 0.12\ast\left[ 0.75+\frac{\left(3.8+1.4Z\right) }{\overline{U }}\right]\)} \\ 
\hline
\multicolumn{1}{|p{1.22cm}}{} & 
\multicolumn{1}{|p{4.61cm}}{Monin-Obukhov length (L)} & 
\multicolumn{1}{|p{2.11cm}}{Parameter} & 
\multicolumn{1}{|p{2.7cm}}{Epistemic} & 
\multicolumn{1}{|p{6.34cm}|}{\(\left\{ L | 10 m\leq L\leq 50 m\right\}\)} \\ 
\hline
\multicolumn{1}{|p{1.22cm}}{} & 
\multicolumn{1}{|p{4.61cm}}{Discretization error} & 
\multicolumn{1}{|p{2.11cm}}{Numerical} & 
\multicolumn{1}{|p{2.7cm}}{Epistemic} & 
\multicolumn{1}{|p{6.34cm}|}{\( Use GCI to characterize\)} \\ 
\hline
\multicolumn{1}{|p{1.22cm}}{} & 
\multicolumn{4}{|p{15.76cm}|}{Source term may not be selective in comparing models} \\ 
\hline
\multicolumn{5}{|p{16.98cm}|}{Propagate Uncertainties} \\ 
\hline
\multicolumn{1}{|p{1.22cm}}{} & 
\multicolumn{4}{|p{15.76cm}|}{\raggedright
Nested uncertainty propagation: aleatory uncertainties in inner loop propagated using progressive LHS; epistemic uncertainties in the outer loop propagated using global optimization methods.} \\ 
\hline
\multicolumn{5}{|p{16.98cm}|}{Present Results} \\ 
\hline
\multicolumn{1}{|p{1.22cm}}{} & 
\multicolumn{4}{|p{15.76cm}|}{Probability box plot and expanded uncertainties} \\ 
\hline
\end{tabular}
\end{adjustbox}
\end{table}

\subsection{Image Classification: MNIST Data}
The image classification task is a core component in machine learning pipelines. In the gas distribution industry, image classification and regression models may be used to identify high consequence areas, for classification of land use and land cover from remote sensing data\citep{Helber.2018}, in the fusion of low-spatial-high-temporal-resolution with high-spatial-low-temporal-resolution data for change and activity detection\citep{Chen.2021}, to detect ground deformation\citep{Anantrasirichai.2020}, and for remote detection of natural gas leaks\citep{Kumar.2020}. Additionally, time series data from field sensors are often converted to images through wavelet decomposition as part of a machine learning model workflow. The subsection will apply the UQ methodology discussed above to the case handwriting image recognition using the well-studied MNIST data\citep{LeCun.2010}.

\subsubsection{MNIST Data}
\begin{figure}
    \centering
    \includegraphics[width=13.2cm]{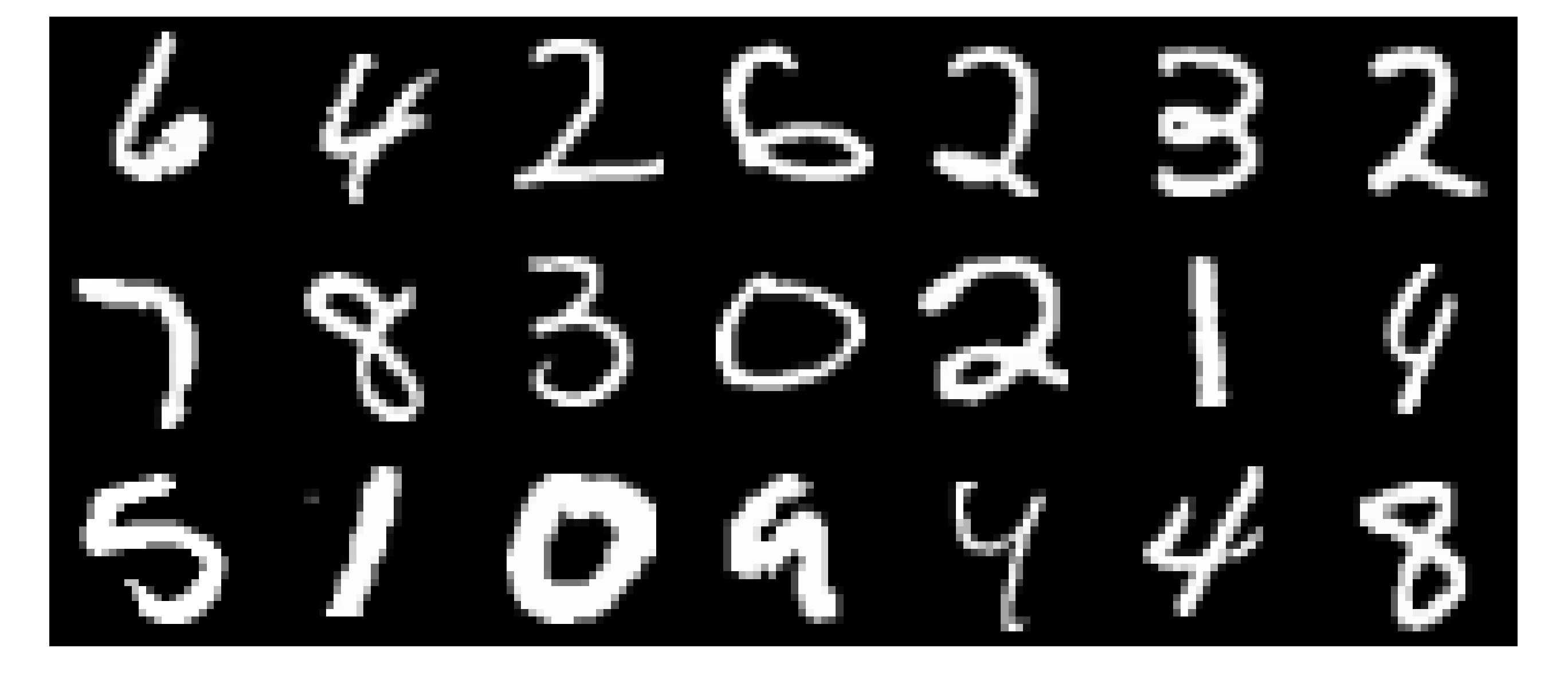}
    \caption{Random sample of digits from the MNIST database.}
    \label{fig:imgMNISTSample}
\end{figure}

The MNIST database is a collection of handwritten digits comprising of labeled 60,000 training examples and 10,000 test examples. Each image is a centered, 28-by-28, 8-bit, grayscale image of handwritten numeral digits, 0 – 9. Samples of the images show a broad range of penmanship styles and pen stroke weights. The images demonstrate varied depictions of the digits that may perplex a machine-learned model. Figure \ref{fig:imgMNISTSample} shows a random sample of images from the MNIST database.

\subsubsection{Classification Model}
A simple two-layer fully connected classification neural network was developed to demonstrate the UQ methodology. The neural network consists of an input layer, two hidden layers, and an output layer. The input layer is of size 784 which corresponds to the vectorized image size. The first hidden layer is a fully connected layer of size 10, with element-wise rectified linear unit (ReLU)  activations. The second hidden layer is a fully connected layer of size 10, corresponding to the size of the output classes. The second hidden layer is followed by elementwise SoftMax activations.

\begin{figure}
    \centering
    \includegraphics[width=13.94cm,height=10.93cm]{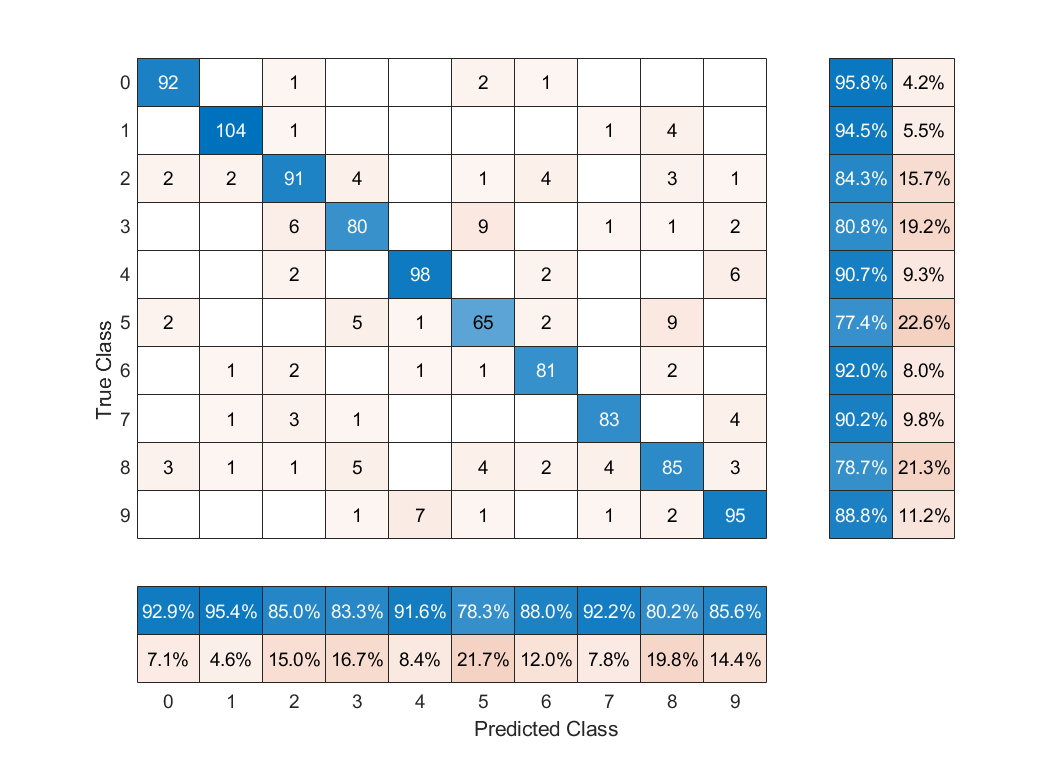}
    \caption{Confusion chart showing performance of baseline classification model. The blue entries in the horizontal bar at the bottom of the figure show the precision of the model for the different classes. The blue entries in the vertical bar show the recall.}
    \label{fig:imgClassficationConfusionChart}
\end{figure}

The model was trained on random selection of 6,000 images from the MNIST training database. The model showed an average validation precision of 0.873 when comparing its predicted labels to ground-truth labels from test set. The model was most predictive with digit classes `0' and `1' and least predictive with digit classes `3', `5', and `8'. The confusion chart shown in Figure \ref{fig:imgClassficationConfusionChart}] shows that that model seems to confuse 3s, 5s, and 8s. Performance of the baseline model is summarized in the precision and recall metrics shown in Figure \ref{fig:imgClassficationConfusionChart}.

\subsubsection{Define Objectives}
Here, we apply the UQ methodology to the case of image classification using the model and data described above. The objective of this UQ study is to assess the predictive capability of the model based on the uncertainties in predictions of the model using samples from the test database. The QOI is the classification of a single sample from the image database. 

\subsubsection{Identify Sources of Uncertainty}
The input to the image classification model is a single 28-by-28, 8-bit, grayscale image. Uncertainty in the input for this model pertains to uncertainty in the image acquisition process. In general, the measurement process (image acquisition in this case) may be abstracted as a convolution of the scene model/process (what is being measured), the sensor model (how it being measured), and the sampling process (how the measured signal is digitized). Uncertainty in the process is introduced in the two latter aspects.

Here, the ``scene" is samples of handwritten digits. The sensor process consists of lighting, an optical system, and an integrated circuit, such as a charge-coupled device. The optical system's response to a point source is spread over a finite spatial area. The sensed intensity at a location is a superposition of spread responses from different point sources. This area is characterized by the width of the point response function (PRF):
\begin{equation}
    d_{s} =  1.22\frac{M\cdot \lambda }{s_{o}\cdot a}
\end{equation}
where \( d_{s}\) is the nominal width of the PRF, \( M\) is the image magnification, \( \lambda\) is the wavelength of the light source, \( s_{o}\) is the distance from the object to the lens, and \( a\) is the effective lens aperture radius\citep{Hecht.1998,Santiago.2001}. This spread may produce a blur in the image if the width of the function is larger than a pixel size. The width of the PRF corresponds to the smallest characteristic length that can be resolved by the optical system. Image processing pipelines use Gaussian filters of sizes same order as the PRF width to remove high frequency noise from images. Samples of the images shown in Figure 5 do not appear to be noisy or blurry. Also, information on the image acquisition system is not readily available. So, input uncertainties are neglected in this analysis. Additional information of characterizing and estimating image noise may be found in these references\citep{Meola.2011,Hytti.2006}. 

Code and solution verification do not apply to machine learning models in the same manner as they do to mechanistic models. The image classification model under analysis does not calculate quantities of interest at discrete temporal and spatial points. So, this model does not incur discretization errors as defined above. The same model used in training is used in simulation/evaluation. The effects of coding errors that may exist in the model are displaced into the inferred model parameters. The effects of floating-point arithmetic are negligible since the model is a classification model. For the UQ study, numerical uncertainties are not a significant source of uncertainties to the computation of the QOI.

Parametric uncertainties are primary sources of uncertainties in the computation of the QOI for this model. Model parameter values are derived from the training data using the learning algorithm. Different samples of the training data will result in different parametrizations. The stability and variance of parameter values is a measure of the uncertainty of the model.

\subsubsection{Characterize Uncertainties}
Parametric uncertainties for the image classification model may be derived by running bootstrap samples of the training dataset through the learning algorithm. These uncertainties are classified as aleatory uncertainties since they are functions of the underlying data distribution.

\subsubsection{Propagate Uncertainties}
\begin{figure}
    \centering
    \includegraphics[height=4in]{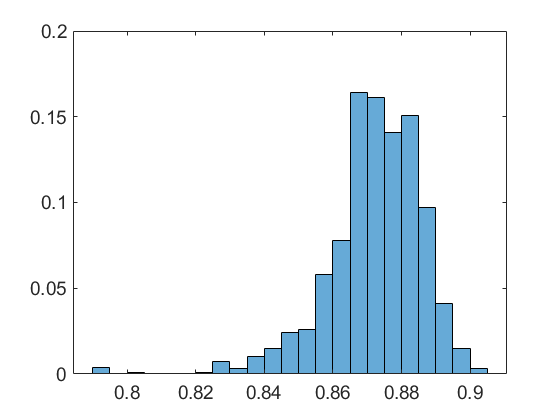}
    \caption{Distribution of validation precision of image classification model for bootstrap size 1000.}
    \label{fig:imgHistPrecision}
\end{figure}

In this example UQ study, a bootstrap size, \( B\), of 1000 and sample size, \( n\), of 6000 were used to generate the posterior distributions of the parameter values. The parameters are the weights and biases of the neural network. The values from the posterior distributions were used in validation tests of the model and the global validation precision for each bootstrapped sample was calculated. Figure \ref{fig:imgHistPrecision} shows the distribution of the average validation precision (over the different digits) of the handwritten digit image classification model. The image shows that the global performance of the classification model ranges between 0.792 and 0.905. While, the mean, median, and mode of the distribution are 0.872, 0.873, and 0.872, respectively. Note that if input uncertainties were not neglected, then a nested uncertainty propagation approach may be used to propagate effects of input and parametric uncertainties.

\subsubsection{Representation of Results}
\begin{figure}
    \centering
    \includegraphics[width=14.33cm,height=18.46cm]{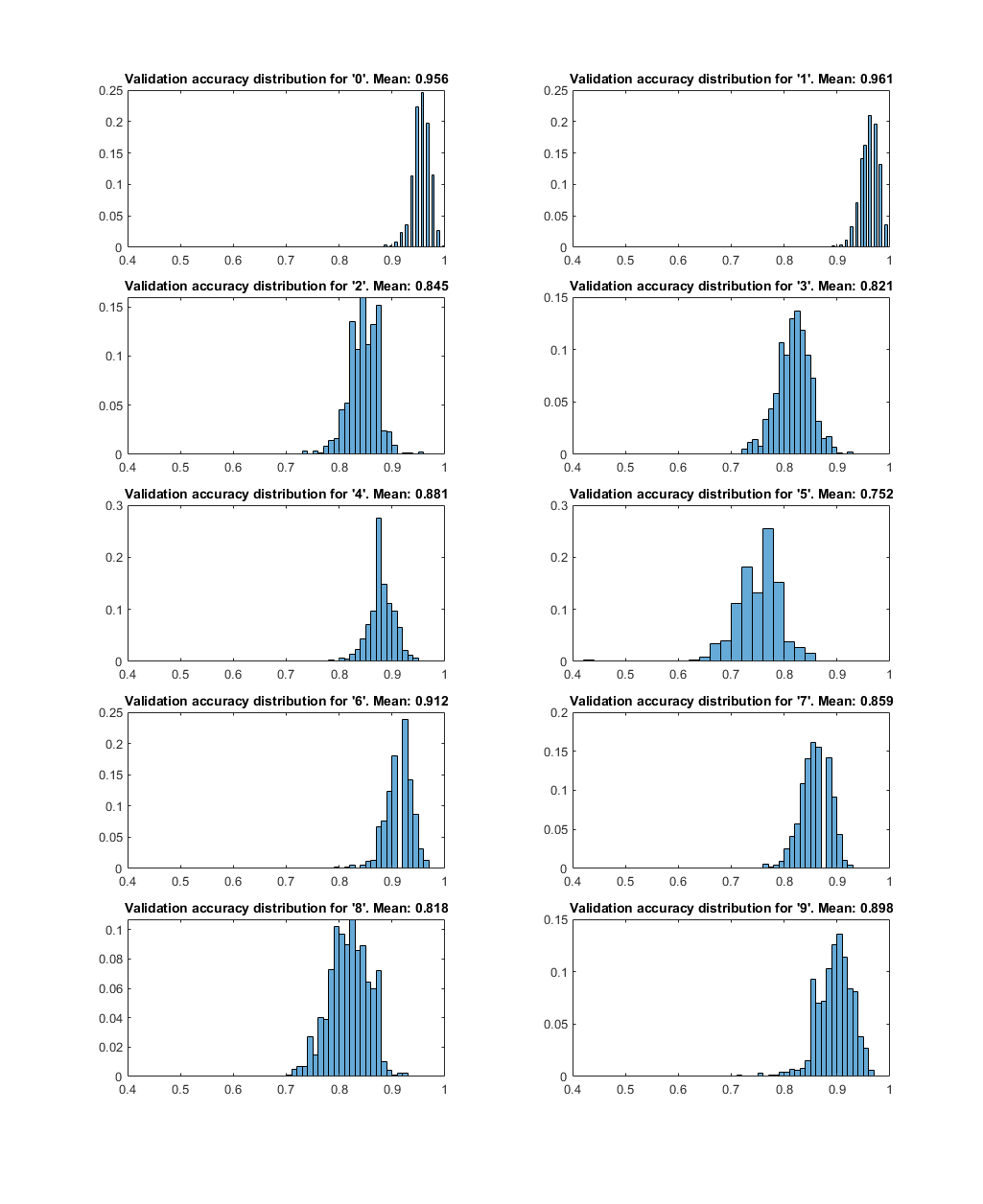}
    \caption{Distribution of validation precisions for image classification of digits 0 -9.}
    \label{fig:imgHistClassificationPrecisions}
\end{figure}

Results from the confusion chart for the baseline model show that predictive capabilities of the model are not similar for all digits. This conclusion is also borne out from the distributions of the validation precisions for the different digits, as shown in Figure \ref{fig:imgHistClassificationPrecisions}. The model may not be suitable for classifying handwritten digits 3, 5, 8 and a better model may be needed.

\begin{table}
\caption{ Summary of handwritten digits classification workflow.}
\label{tab:summaryOfDigitsClassificationWorkflow}
\begin{adjustbox}{max width=\textwidth}
\begin{tabular}{p{1.22cm}p{4.61cm}p{2.11cm}p{2.7cm}p{6.34cm}}
\hline
\multicolumn{5}{|p{16.98cm}|}{Define QOI} \\ 
\hline
\multicolumn{1}{|p{1.22cm}}{} & 
\multicolumn{4}{|p{15.76cm}|}{\raggedright
Classification of a single image sample} \\ 
\hline
\multicolumn{5}{|p{16.98cm}|}{Identify Sources of Uncertainties} \\ 
\hline
\multicolumn{1}{|p{1.22cm}}{} & 
\multicolumn{4}{|p{15.76cm}|}{\raggedright
Parametric uncertainties, input uncertainties.} \\ 
\hline
\multicolumn{5}{|p{16.98cm}|}{Characterize Uncertainties} \\ 
\hline
\multicolumn{1}{|p{1.22cm}}{} & 
\multicolumn{1}{|p{4.61cm}}{\centering
\textbf{Uncertain Parameter}} & 
\multicolumn{1}{|p{2.11cm}}{\centering
\textbf{Category}} & 
\multicolumn{1}{|p{2.7cm}}{\centering
\textbf{Classification}} & 
\multicolumn{1}{|p{6.34cm}|}{\centering
\textbf{Characterization}} \\ 
\hline
\multicolumn{1}{|p{1.22cm}}{} & 
\multicolumn{1}{|p{4.61cm}}{\raggedright
Neural network weights and biases} & 
\multicolumn{1}{|p{2.11cm}}{Parameter} & 
\multicolumn{1}{|p{2.7cm}}{Aleatory} & 
\multicolumn{1}{|p{6.34cm}|}{Use bootstrap to characterize} \\ 
\hline
\multicolumn{1}{|p{1.22cm}}{} & 
\multicolumn{4}{|p{15.76cm}|}{Input uncertainties were neglected because images do not appear noisy or blurry, and because details of the image acquisition process in unknown} \\ 
\hline
\multicolumn{5}{|p{16.98cm}|}{Propagate Uncertainties} \\ 
\hline
\multicolumn{1}{|p{1.22cm}}{} & 
\multicolumn{4}{|p{15.76cm}|}{\raggedright
Monte Carlo sampling} \\ 
\hline
\multicolumn{5}{|p{16.98cm}|}{Present Results} \\ 
\hline
\multicolumn{1}{|p{1.22cm}}{} & 
\multicolumn{4}{|p{15.76cm}|}{Probability distribution plots} \\ 
\hline
\end{tabular}
\end{adjustbox}
\end{table}

\section{Conclusions}
This project developed an uncertainty quantification protocol with which to assess the fitness-for-use of machine learning and mechanistic models. This reports details backgrounds of different statistical methods and their usefulness to uncertainty quantification of models of interest. This report serves as a reference from which specific UQ workflows may be developed. 

\subsection{Integrated Workflow}
\label{subsec:integratedWorkflow}
The DAKOTA software\citep{Bohnhoff.2019uf}is a freely-available, general-purpose, state-of-the-art, toolkit for optimization, parameter estimation, sensitivity analysis, and uncertainty quantification. The software, developed by Sandia National Laboratory, contains a full suite of mathematical and statistical methods, and computational algorithms to execute all facets of the UQ protocol discussed in this report. The toolkit has flexible and extensible interface that facilitates connections to external simulation codes. The toolkit also has direct integrated connectivity to common modeling platforms such as MATLAB and Python. DAKOTA comes with a graphical user interface (GUI) process inputs and outputs from simulation studies. DAKOTA is a complete tool where a non-expert in UQ may apply the protocol developed in this project to assess the credibility of machine learning and mechanistic models.

The workflows highlighted in the preceding section address common classes of models employed in the gas distribution industry. Probabilistic models, such Bayesian networks, are another class of commonly used models in the industry. Using the principles discussed in this protocol, templates for these classes of models may be developed for UQ studies in the DAKOTA framework. This will reduce the barrier to conducting UQ studies and increase the adoption of conducting uncertainty quantification studies in the industry.

\newpage
\bibliographystyle{unsrtnat}
\bibliography{VVUQ_Nat_Gas}

\end{document}